# Synthetic Data Generation for Emotional Depth Faces: Optimizing Conditional DCGANs via Genetic Algorithms in the Latent Space and Stabilizing Training with Knowledge Distillation

Seyed Muhammad Hossein Mousavi
C.E.O. and Researcher at Cyrus Intelligence Research Ltd
Tehran, Iran
ORCID: 0000-0001-6906-2152
s.m.hossein.mousavi@cyrusai.ir

S. Younes Mirinezhad
Computer Vision Technical Lead at Pars AI
Tehran, Iran
ORCID: 0000-0003-4723-3322
younes.mirinezhad@gmail.com

*Abstract*

Affective computing faces a significant challenge: the scarcity of high-quality, diverse depth facial datasets crucial for understanding subtle emotional expressions. This data gap prevents the development of robust models for applications ranging from human-computer interaction to healthcare. To address this, we propose a novel framework for synthetic depth face generation using an optimized Generative Adversarial Network (GAN). First, we integrate Knowledge Distillation (KD), using Exponential Moving Average (EMA) teacher models to stabilize GAN training, enhance the quality and diversity of generated images, and help prevent mode collapse. More importantly, we extend the use of Genetic Algorithms (GAs) into the GAN's latent space. By evolving these latent vectors based on image statistics (standard deviation and mean) that correlate with visual quality and diversity, the GA helps to identify latent representations that yield high-quality and varied images for a given target emotion, thereby optimizing the visual output for specific emotional categories. Thus, GA acts as post-processing, which diversifies faces even more than the KD level. This comprehensive methodology not only mitigates data scarcity but also yields diverse images and promising results compared to other benchmark techniques such as GAN, VAE, Gaussian Mixture Models (GMM), and Kernel Density Estimation (KDE). For the machine learning aspect, we extract depth-related LBP, HOG, Sobel edge histogram, and intensity histogram features and concatenate them for the classification task. Metrics responsible for comparing synthetic and original samples, plus evaluating synthetic samples in terms of quality and diversity, include Fréchet Inception Distance (FID), Inception Score (IS), Structural Similarity Index Measure (SSIM), and Peak Signal-to-Noise Ratio (PSNR). Our method achieved 94% and 96% recognition accuracy with the XGBoost classifier and surpassed other state-of-the-art methods. Furthermore, our methods outperform other methods regarding all synthetic data evaluation metrics.

*Keywords*: Emotion Recognition, Data Scarcity, Depth Images, Facial Expressions, Optimization, Genetic Algorithm, Knowledge Distillation

- **Introduction**

Emotion Recognition (ER) is a cornerstone of affective computing [1], a rapidly expanding field dedicated to enabling machines to understand, interpret, and even express human emotions. Its importance spans diverse applications, from creating more empathetic human-computer interactions and personalizing educational experiences to enhancing mental health monitoring and developing advanced robotics. By allowing technology to perceive emotional cues, we move closer to a more intuitive and human-centric digital world. However, building robust emotion recognition systems is fraught with challenges. Human emotions are inherently complex, nuanced, and often expressed subtly [2], making their accurate detection difficult. Furthermore, variations in individual expression, cultural context, and environmental factors add layers of complexity to the task [3].

A significant hurdle in training effective deep learning models for emotion recognition is data scarcity [4-6]. High-quality, diverse, and well-annotated datasets are the lifeblood of modern AI, but they are often expensive, time-consuming, and difficult to acquire. This limitation severely impacts the ability of models to generalize well to unseen data and truly capture the full spectrum of human emotional expression [7]. This challenge is particularly acute when it comes to facial images, especially depth facial images [4, 5]. While 2D facial images have been widely studied, depth information provides crucial 3-Dimensional (3D) structural cues that can offer a richer, more robust understanding of facial expressions, being less susceptible to lighting variations or occlusions. Yet, obtaining large, varied datasets of depth faces across different emotional states is exceptionally difficult due to specialized hardware requirements, privacy concerns, and the sheer effort involved in data collection and annotation. This critical lack of emotionally rich, depth facial data directly impedes the development of next-generation affective computing systems that could leverage these valuable 3D insights [8].

Despite recent advances in deep learning and affective computing, the lack of large-scale, diverse, and emotion-specific depth facial datasets remains a fundamental bottleneck. Unlike RGB data, depth information captures subtle geometric and structural cues that are essential for recognizing fine-grained emotional expressions, especially in real-world, unconstrained environments. However, collecting such datasets is expensive, time-consuming, and often limited by ethical and privacy concerns. This motivates the need for a synthetic data generation framework that can produce high-quality, diverse, and emotionally relevant depth faces. Our work aims to fill this gap by using generative models, optimization algorithms, and knowledge transfer strategies to simulate realistic emotional depth data for downstream classification and analysis.

In this study, we also propose an integrated emotional representation model that synthesizes elements from Lisa Feldman Barrett's theory of constructed emotion [9] and James Russell's circumplex model of affect[1] [10] in Figure 1. We proposed this model as we believed other available models either lack a number of emotions to cover all human daily life emotion experiences, or they are not accurate enough. Black dots belong to Russell's model, and Blue ones belong to Feldman's model. Yellow indicators are connections of emotions that bring engagement or activity in educational systems. Emotions with red rectangles around them indicate their relevance and importance, with the yellow indicators. Our model uses Russell's core dimensions of Valence (ranging from unpleasant to pleasant) and Arousal (ranging from low to high) to spatially position a comprehensive range of emotional states. This dimensional structure is enriched with conceptual contributions from Feldman Barrett's theory, which emphasizes the role of contextual interpretation, conceptual knowledge, and internal bodily states in the dynamic construction of emotion. By overlaying these perspectives, our model enables a more subtle classification of emotional states, such as anxious, frustrated, or active/engaged, based not only on physiological arousal and hedonic tone but also on inferred skill levels and cognitive appraisal cues. The inclusion of labeled quadrants (tension, energy, tiredness, calmness) supports interpretability in affective computing systems, making this model particularly well-suited for applications involving facial expressions, depth video analysis, and multimodal emotion recognition in both controlled and real-world settings.

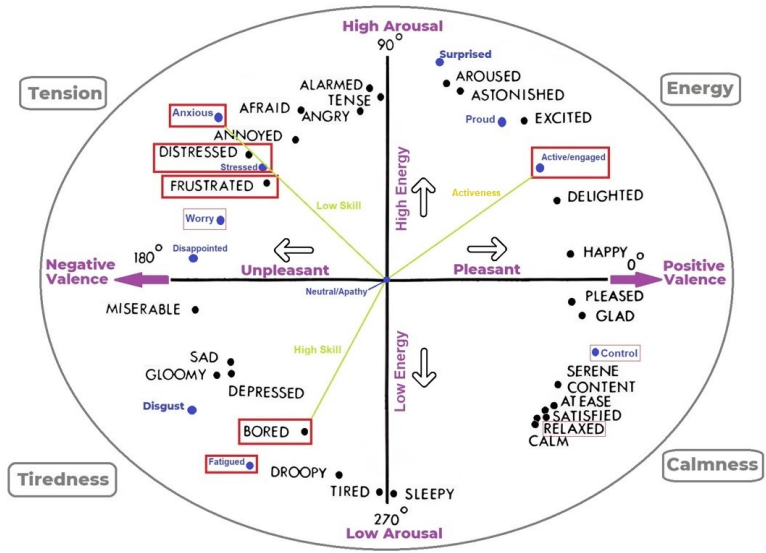

Figure 1. Our proposed emotion recognition model

➢ **Synthetic Data Generation (SDG)**

Generative AI is a branch of artificial intelligence that focuses on learning data patterns to produce new, realistic content across multiple formats [11]. Unlike discriminative AI, which maps data to labels (as in classification tasks), generative AI works in reverse, starting from labels to generate diverse forms of data. This capability underpins various applications, such as image creation, text generation, text-to-image synthesis, music composition, video production, digital artwork, and transforming textual prompts into audio or visual content. One of its key uses is in the creation of synthetic data, a process formally known as Synthetic Data Generation (SDG) [6,7,12-15]. SDG presents a compelling solution to issues like limited data availability by enabling the production of vast, varied, and controlled datasets. These datasets can significantly boost the training and accuracy of machine learning models, particularly in emotion recognition. Moreover, SDG allows for the modeling of subtle emotional states that are often underrepresented in real-world datasets. Through algorithms that emulate authentic human emotions and behaviors, researchers can generate lifelike data suited for building more resilient and precise emotion recognition systems. Additionally, synthetic data contributes to more ethical AI development by minimizing the need for real personal data, thereby addressing privacy concerns and playing a foundational role in future AI advancements.

---

[1] https://www.sciencedirect.com/topics/psychology/circumplex-model

> **Depth Image Data Type**

A depth image, also known as a depth map, is a type of image data where each pixel represents the distance from the sensor to the nearest surface in the scene, rather than color or intensity information [16]. This modality provides three-dimensional structural information by encoding depth values, typically in millimeters, using devices such as the Microsoft Kinect sensor[2]. Depth images are crucial in many computer vision applications because they offer valuable spatial context that RGB images alone cannot provide. They play an essential role in robotics, augmented reality, gesture recognition, and human-computer interaction due to their ability to capture precise geometric details. In recent years, depth images have gained significant attention in Facial Expression Recognition (FER) and emotion recognition, as the 3D shape of facial muscles and subtle movements are better captured through depth data than traditional 2D imaging. This enhanced capability allows for more robust emotion classification, especially in varying lighting conditions or complex backgrounds. The key feature of depth images lies in the fact that each pixel corresponds to the distance between the subject and the depth sensor, often measured in millimeters when using time-of-flight or structured light sensors like the Kinect [2, 16]. This per-pixel distance mapping allows for accurate modeling of facial surfaces, making depth data a powerful modality for analyzing and interpreting human emotions. Figure 2 depicts a sample from the IKFDB dataset [16] in grayscale and depth format, which represents different forms of pixel values for calculation. Darker depth pixels represent a closer distance to the sensor.

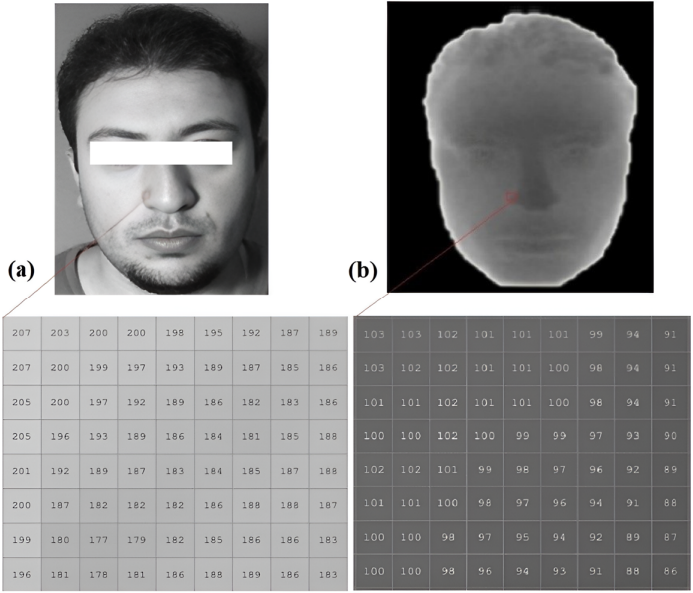

Figure 2. A selected sample from the IKFDB dataset in grayscale and depth format, representing their difference based on pixel values

> **Conditional Deep Convolutional Generative Adversarial Network (cDCGAN)**

A Conditional Deep Convolutional Generative Adversarial Network (cDCGAN) [17] is an extension of the standard Deep Convolutional GAN (DCGAN) [17] architecture, which combines Convolutional Neural Networks (CNN) [18] with the generative power of GANs and introduces conditional inputs to control the generation process. In a typical GAN [11] setup, a generator creates synthetic data while a discriminator attempts to distinguish between real and generated data. The conditional variant enhances this framework by incorporating additional information, such as class labels, attributes, or other context, into both the generator and discriminator, guiding the generation process toward more meaningful and controllable outputs. In cDCGAN, the conditioning data helps the generator produce class-specific or attribute-specific samples, while the discriminator becomes better at distinguishing between real and synthetic data conditioned on the same information. This makes cDCGAN particularly powerful for tasks requiring structured output and control, such as data augmentation and image synthesis. In the context of depth images, especially for emotion recognition, cDCGAN has shown great promise in generating realistic synthetic depth maps of facial expressions corresponding to specific emotional states. This is particularly valuable for overcoming data scarcity in emotion recognition datasets, where obtaining large-scale, annotated depth image data is challenging. By generating diverse and emotion-specific depth facial samples, cDCGAN can enrich training datasets, improve model generalization, and enhance the performance of emotion recognition systems that rely on depth-based facial geometry and spatial cues. Figure 3 illustrates the flowchart of the cDCGAN algorithm. We use cDCGAN to generate synthetic depth facial images conditioned on specific emotions, allowing the model to learn emotion-specific features and produce targeted, high-quality samples. This directly addresses the lack of diverse emotional depth data crucial for affective computing applications.

---
[2] https://azure.microsoft.com/en-us/products/kinect-dk

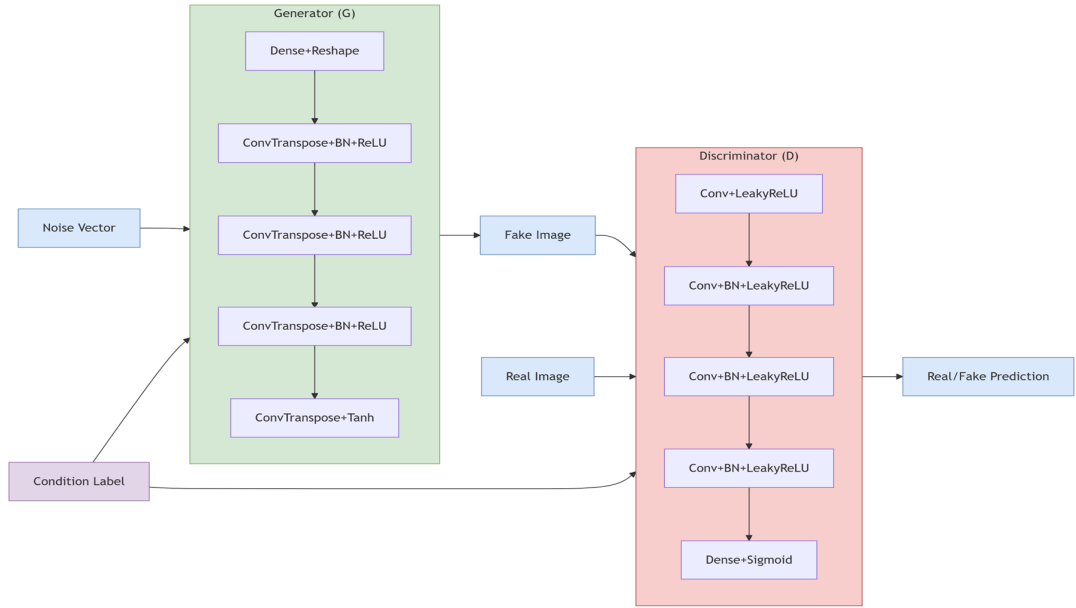

Figure 3. The cDCGAN algorithm's flowchart

➢ **Optimization and Genetic Algorithm (GA)**

Optimization algorithms are mathematical methods used to find the best possible solution, often under a set of constraints, by maximizing or minimizing an objective function. They play a crucial role in a wide range of fields, from engineering design and economics to machine learning and artificial intelligence, where finding optimal parameters, structures, or decisions is vital for performance and efficiency [19]. These algorithms are widely applied in logistics for route planning, in finance for portfolio optimization, in healthcare for treatment planning, and in AI for model training and hyperparameter tuning. However, optimization problems often present significant challenges such as non-linearity, high dimensionality, multiple local optima, and noisy or incomplete data, which make traditional methods less effective in complex real-world scenarios [20, 21]. Among the various optimization techniques, Genetic Algorithms (GA) [22] stand out as one of the most powerful and flexible approaches. Inspired by the process of natural selection and evolution, a Genetic Algorithm is a population-based metaheuristic that evolves a set of candidate solutions using operations such as selection, crossover (recombination), and mutation. The importance of GA lies in their ability to efficiently explore large and complex search spaces without requiring gradient information or precise mathematical formulations, making them especially suitable for black-box or multi-modal problems. GA has been successfully applied in numerous domains, including feature selection in machine learning, scheduling and resource allocation, robotics path planning, neural network architecture optimization, and even in bioinformatics for protein structure prediction. Their adaptability and robustness continue to make them a valuable tool for solving a wide variety of optimization problems where traditional approaches may fall short. Figure 4 depicts the flowchart of the genetic algorithm. We use GA to optimize the latent space of the cDCGAN, enabling more stable training and generation of high-quality, diverse depth facial images. This approach addresses data scarcity by evolving emotionally relevant representations that outperform traditional methods in quality and diversity.

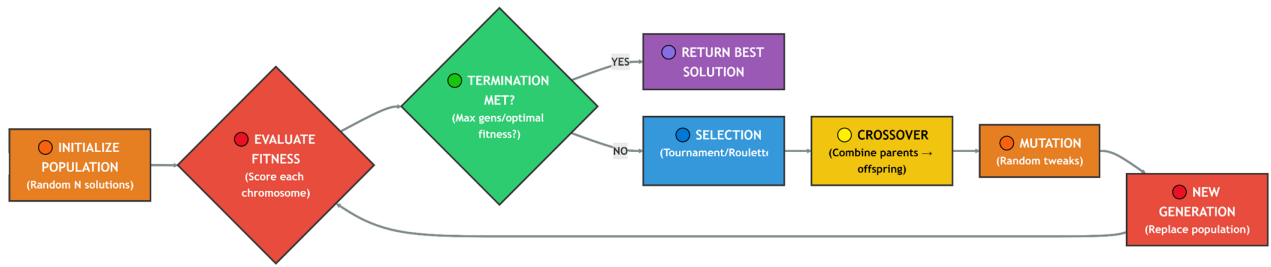

Figure 4. The flowchart of the genetic algorithm

> **Knowledge Distillation (KD)**

Knowledge Distillation (KD) [23] is a model compression technique in machine learning where a smaller, simpler model (called the student) is trained to replicate the behavior of a larger, more complex model (called the teacher). Instead of learning directly from raw data, the student model learns from the softened output probabilities or feature representations generated by the teacher, which contain richer information than hard labels alone. This approach is important because it enables the deployment of high-performance models in resource-constrained environments such as mobile devices, embedded systems, and edge computing platforms, where storage, computation, and energy are limited. Knowledge distillation has found wide applications in fields like image classification, natural language processing (e.g., compressing large transformers like BERT), speech recognition, and even reinforcement learning, where maintaining model performance while reducing complexity is crucial. Despite its advantages, knowledge distillation faces several challenges, such as how to effectively transfer knowledge when the student and teacher architectures differ significantly, choosing the right temperature and loss functions, and validating that the distilled knowledge does not lead to overfitting or performance degradation in unseen data. We use KD to stabilize GAN training and enhance the quality and diversity of generated depth facial images by transferring refined knowledge from an Exponential Moving Average (EMA)[3] [24] teacher model. This helps prevent mode collapse and improves the generation of emotion-specific facial data. Figure 5 illustrates the KD's flowchart.

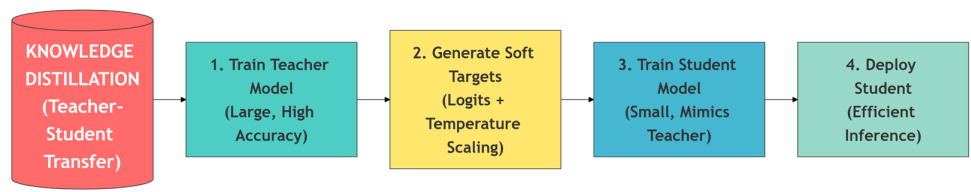

Figure 5. The flowchart of the knowledge distillation training strategy

> **Contribution**

In this work, we present a comprehensive and novel contribution to the field of affective computing by addressing the critical problem of data scarcity in depth-based facial emotion datasets. Our main contribution lies in the development of an end-to-end framework that synthesizes high-quality and emotion-specific depth facial images, empowering downstream emotion recognition tasks. At the core of our method is an optimized cDCGAN, tailored to generate synthetic depth images conditioned on emotional categories. To enhance the training efficiency and performance of the cDCGAN, we employ GA for evolving latent vectors based on fitness functions derived from statistical image quality metrics such as standard deviation and mean intensity. This evolutionary process discovers optimal latent codes that produce diverse and high-quality facial expressions aligned with target emotional labels, effectively acting as a post-generation enhancement mechanism. Furthermore, we incorporate KD by introducing an EMA teacher model during training. This helps stabilize GAN learning, mitigate mode collapse, and guide the generator towards producing more consistent and semantically meaningful outputs. The interplay between KD and GA yields a hybrid optimization-learning strategy that not only boosts generation fidelity but also maintains emotional distinctiveness across samples. To validate the efficacy of our method, we extract and concatenate handcrafted features such as Local Binary Patterns (LBP) [25], Histogram of Oriented Gradients (HOG) [26], Sobel edge histograms [27], and intensity histograms [27] from the generated images and use them for emotion classification. Our synthetic data shows high utility and realism, as evidenced by both visual inspection and quantitative evaluation using Fréchet Inception Distance (FID) [28], Inception Score (IS) [29], Structural Similarity Index Measure (SSIM) [30], and Peak Signal-to-Noise Ratio (PSNR) [31]. In comparison to conventional generative models, including standard GAN [11], Variational Auto Encoders (VAE) [32], Gaussian Mixture Models (GMM) [33], and Kernel Density Estimation (KDE) [34], our approach exhibits superior performance in terms of both diversity and emotional specificity. In summary, our contribution is a unique multi-stage pipeline that combines the strengths of evolutionary optimization, knowledge transfer, and conditional generation to address the crucial issue of emotional depth image synthesis for affective computing applications.

---

[3] https://medium.com/analytics-vidhya/understanding-exponential-moving-averages-e3f020d9d13b

> ➢ **Research Questions**

Given the increasing demand for rich, diverse, and high-quality depth facial data in affective computing, we first ask: (1) How can synthetic depth face generation effectively address the scarcity of emotion-specific datasets? Building on this, we consider (2) Can KD, particularly using an EMA teacher model, contribute to improved GAN training by reducing mode collapse and enhancing image consistency? As diversity is key in emotion synthesis, we then ask: (3) How can GA be extended into the latent space to evolve vectors that produce high-quality and emotionally diverse depth faces? From a classification standpoint, a practical concern is (4) Do features extracted from synthetic depth images (LBP, HOG, Sobel edges) preserve enough emotional information to perform reliably in recognition tasks? Finally, to validate generative success beyond visual inspection, we inquire: (5) How do quantitative metrics such as FID, IS, SSIM, and PSNR reflect the visual quality and emotional diversity of the generated samples compared to other generative approaches like VAE, GMM, and KDE?

> ➢ **Paper Structure**

The structure of this paper is designed to guide the reader from foundational knowledge to detailed implementation and analysis. We begin with the introduction, where we outline all the main definitions, motivation behind our work, highlight the challenges in affective computing, particularly the scarcity of depth-based emotional datasets, and briefly present our proposed solution. This is followed by the prior related research section, which reviews state-of-the-art SDG algorithms and explores relevant literature across various modalities, including physiological signals, body motion, sound, and particularly color and depth images in emotion recognition. The proposed method section then outlines our theoretical approach and implementation details. In the evaluation and results section, we compare our method against existing depth image synthesis techniques, describe the dataset used, detail the feature extraction process, evaluation metrics, classifiers, and report experimental findings along with a critical discussion. Finally, the conclusion summarizes key contributions, with dedicated subsections for suggestions and future works to highlight potential directions for expanding and improving this research.

- **Prior Related Research**
  > ➢ **SDG State of the Art (SoA) Algorithms**

In the domain of SDG, algorithms are generally classified into non-learning-based (unintelligent) and learning-based (intelligent) techniques. Non-learning-based methods typically include conventional data augmentation techniques such as image rotation, scaling, and filtering. Among statistical methods, Principal Component Analysis (PCA) is often applied to reduce data dimensionality by identifying key features that capture the majority of variance [35]. Another widely used method, Synthetic Minority Over-sampling Technique (SMOTE), generates synthetic examples to address class imbalance issues in classification tasks [36]. The GMM performs data augmentation by sampling new synthetic data points from the learned distribution of the original dataset, capturing complex multimodal patterns. Each Gaussian component represents a cluster, and new points are generated by sampling from these components [64]. Also, the KDE performs data augmentation by estimating the probability density function of the original data and then sampling new synthetic points from this smoothed distribution, preserving the structure of the dataset without assuming a parametric form [65]. On the other hand, intelligent methods, which rely on machine learning models, have seen broader adoption in recent years. Prominent among them are Generative Adversarial Networks (GAN) [11], which train two competing networks to generate realistic synthetic data, and Variational Autoencoders (VAE) [32], which learn probabilistic latent representations for generative purposes. Models like Long Short-Term Memory (LSTM) networks are valuable for sequential data synthesis due to their memory-retaining capabilities [37], while copula-based models offer flexibility in modeling dependencies between variables for multivariate data simulation [38]. Recent advancements have introduced powerful deep learning-based generators such as diffusion models [33] and transformer architectures [34], both known for their high-quality generative performance across modalities. Additionally, Convolutional Neural Networks (CNN) remain effective for generating image-like data due to their hierarchical spatial feature extraction abilities [18]. Figure 6 represents the SDG main algorithms.

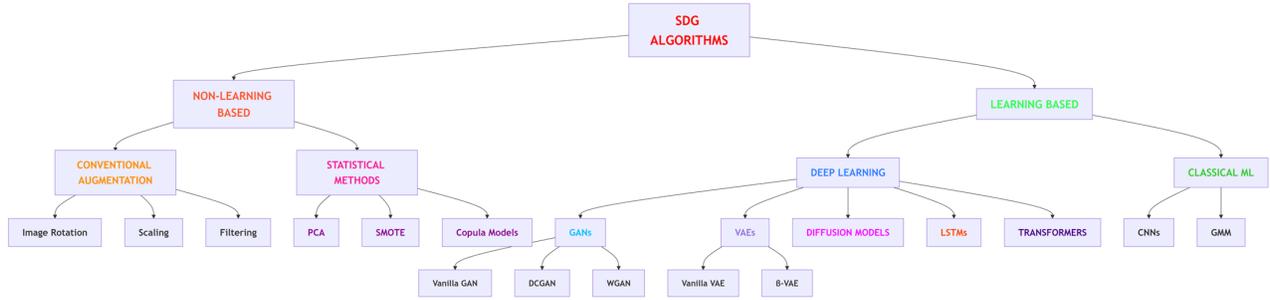

Figure 6. Synthetic data generation main algorithms

> **SDG Literature on Non-Image Modalities**

The following research [39] explores the application of Borderline-SMOTE for augmenting Electro Encephalo Graphy (EEG) data to improve emotion recognition using convolutional neural networks. The research focuses on enhancing data quality and model performance in the domain of brain-computer interfaces for emotion recognition. Also, [40] uses PCA for synthesizing tabular data similar to the probabilistic physiological data type. The following research [41] used NGN for SDG of emotional and physiological data (EEG, Electro Cardio Graphy (ECG), and Galvanic Skin Response (GSR)), increasing synthesis quality and optimizing runtime speed. Implementation can be found here[4]. Also, the same author used the NGN for synthesizing the body motion modality in emotion recognition[5] [6]. The [66] proposes a novel GMM-based data augmentation method for EEG signals to improve Brain-Computer Interface (BCI) emotion recognition. By decomposing EEG data into Gaussian components and swapping same-class segments across trials, the approach synthesizes new samples that preserve the temporal-spatial dynamics of the brain signals. Also, the [34] introduces a KDE-based sampling method to address class imbalance in tabular data, which is a common challenge in non-image domains like physiological monitoring. The method estimates the probability density of minority class distributions using KDE and generates synthetic points by sampling from this smooth distribution. It consistently outperforms traditional oversampling techniques such as SMOTE across decision trees, neural networks, and SVM classifiers. The research [42] focuses on generating texts with specific sentiment/emotion labels. This method addresses challenges like the poor quality and lack of diversity in generated texts by implementing a mixture of adversarial networks, potentially improving user interaction systems in educational and therapeutic settings. For audio modality, the [43] combines a VAE with a GAN to perform emotional speech conversion. The VAE component is used to encode emotional nuances from speech, which are then enhanced for clarity and realism by the GAN component, synthesizing new vocal expressions. The [44] presents a method for generating synthetic cardiac signals by employing a copula-based statistical approach. The authors utilize marginal distributions of key parameters and their linear correlations to define a Gaussian copula. From this copula, random sets of parameters are generated, which are then used to synthesize cardiac signals. The research introduced in [45] integrates transformer models into a cycle GAN framework to enhance emotional voice conversion. The study investigates the transformer's ability to capture intra-relations among frames by augmenting the receptive field of models, aiming to improve the conversion of emotional expressions in speech. Also, for SDG using CNN, the authors of [46] used the CNN algorithm for synthesizing emotional ECG signals on the DREAMER database[6], generating a realistic and diverse dataset, and answering data scarcity challenges. The [47] introduces a GAN-based technique for augmenting motion capture datasets with synthetic data. Another example that uses GAN for motion synthesis is [48], which introduces a novel framework for synthesizing diverse and high-quality human motions from a limited set of examples. Also, examples of motion synthesis using Conditional Variational Auto-Encoder (CVAE) are introduced in [49, 50]. The framework in [49] addresses various tasks such as motion prediction, completion, interpolation, and spatial-temporal recovery by treating inputs as masked motion series. It estimates parametric distributions from these inputs to generate full motion sequences and incorporates Action-Adaptive Modulation (AAM) to adjust motion styles based on action labels, enhancing the realism and coherence of the synthesized motions.

> **SDG Literature on Color and Depth Image Modalities in Emotion Recognition**

The [67] present a framework for 3D face recognition using region covariance descriptors extracted from depth scans, modeled with GMM. They fuse multiple local facial surface descriptors (shape index, Gaussian curvature, normals, etc.) into covariance matrices, transform to a compact feature space, and then fit a GMM to capture multimodal depth-based facial variation. Adams et al. In [68], they employ KDE in the embedded latent space of 3D facial shapes, setting Gaussian kernels around each shape point. New synthetic embeddings are drawn from this estimated density and then matched to the closest real shapes to build realistic augmented depth faces. This augmentation mimics expression-like depth variations without using explicit expression labels. The following research [51] employs an LSTM to extract emotions from text, which then guides a GAN to modify facial expressions in images accordingly. The LSTM processes the textual input to determine the desired emotional expression, and the GAN alters the facial image to match this emotion. This study [52] addresses the task of altering facial expressions in images using conditional diffusion models. The authors propose incorporating a semantic encoder to guide the diffusion process, enabling the generation of specific emotional expressions while preserving the individual's identity. In the following research [53], they introduced a dual-GAN framework that learns from RGB and dense depth (plus surface normals), imposing depth-consistency and confidence regularizer losses to ensure geometric plausibility. They reconstruct depth maps from AffectNet and RaFD images via 3D modeling to build RGB–Depth training pairs, enabling superior photorealistic expression generation. The method significantly outperforms earlier baselines on both AffectNet-D and RaFD-D benchmarks. Also, at [54], they rendered thousands of synthetic RGB–Depth facial sequences with controlled variations in expression, pose, lighting, and camera viewpoint, supplying CNNs with precise ground-truth depth for training. Their CNN model achieves robust monocular depth estimation on real facial data, even under varied expressions, and shows promise in applications like in-cabin driver monitoring and robotics. At [55], they proposed a denoising diffusion probabilistic model that learns to generate dynamic 3D facial expression sequences (i.e., 4D faces) by first generating landmark trajectories and then driving mesh deformation through a decoder. This two-stage DDPM framework captures realistic temporal transitions in depth-based facial expressions and generalizes to arbitrary 3D face meshes. Researchers of [56] synthesized over 600k depth-annotated frames using 3D morphable models fitted to 4DFAB with valence–arousal labels, densely aligning face meshes via UV mapping and thin-plate spline warping. By generating expressions across the VA space, they enhanced training data diversity, leading to superior FER performance compared to standard GAN methods. In [57], researchers developed a framework combining text-driven and action-unit GAN controls to generate synthetic facial expression images at scale, introducing semantic guidance and pseudo-labeling to ensure label reliability. Although this work focuses on RGB, its AU-based design is easily extendable to depth data, offering fine-grained control of geometry for depth-aware FER. Paper [58] modeled identity-agnostic expression change by learning feature-level differences between neutral and emotional expressions, enabling the generation of identity-preserving facial expressions with high realism. The conditional autoencoder structure supports interpolation and transfer of expressions, which can be adapted for depth map synthesis by encoding geometric deltas. The paper [59] introduced FESGAN, a unified framework that simultaneously trains a generative model to synthesize varied expressions (including for new identities) and a recognition network, connected through intra-class loss and real-data guided backpropagation to reduce domain shifts. This joint learning strategy can be applied to depth channels to bridge real-synthetic gaps for depth-based FER.

---

[4] https://github.com/SeyedMuhammadHosseinMousavi/Synthetic-Data-Generation-by-Supervised-Neural-Gas-Network
[5] https://github.com/SeyedMuhammadHosseinMousavi/Synthetic-Data-Generation-of-Body-Motion-Data-by-Neural-Gas-Network-for-Emotion-Recognition
[6] https://zenodo.org/records/546113

- **The Proposed Method**

We start with a conditional GAN framework where both the generator $G$ and the discriminator $D$ are conditioned on an emotion label $y \in \{1, \dots, C\}$, and latent vector $z \sim \mathcal{N}(0, I)$.

Equation (1): cDCGAN loss function

$$\mathcal{L}_{\text{cDCGAN}} = \mathbb{E}_{x,y}[\log D(x,y)] + \mathbb{E}_{z,y}[\log(1 - D(G(z,y),y))] \tag{1}$$

- ✓ $G(z, y)$ : generated image
- ✓ $x$ : real depth image
- ✓ $y$ : emotion label
- ✓ $D(x, y)$ : discriminator output
- ✓ This forms the foundation of the model.

After initial training and in the second step, we further optimize the latent vector $z$ using a second GA stage to generate higher-quality images with stronger emotional traits.

Define a fitness function over generated images based on image statistics (mean intensity $\mu$ and standard deviation $\sigma$):

Equation (2): Fitness function for latent vector evolution

$$\mathcal{F}_{\text{quality}}(z) = \alpha \cdot \mu(G(z,y)) + \beta \cdot \sigma(G(z,y)) \tag{2}$$

Where:

- $\alpha, \beta$ are weighting factors
- $\mu(G(z,y))$ : average intensity of the generated image
- $\sigma(G(z,y))$ : spread/contrast

Now we evolve $z$ via GA to maximize this fitness:

Equation (3): GA optimization in latent space

$$z^* = \arg \max_{z \in Z} \mathcal{F}_{\text{quality}}(z) \tag{3}$$

This produces diverse and emotionally aligned images from the generator $G$ already trained.

Additionally, during GAN training, to stabilize learning and prevent mode collapse, we maintain an EMA-based teacher generator $G_t$, updated as:

Equation (4): EMA teacher update

$$G_t^{(t)} = \tau \cdot G_t^{(t-1)} + (1 - \tau) \cdot G_s^{(t)} \tag{4}$$

Where:
- $G_s$ : current student generator (same $G$ from Equation 1 )
- $\tau$ : decay factor (e.g., 0.999)

Then, enforce consistency between teacher and student using L2 loss:

Equation (5): KD consistency loss

$$\mathcal{L}_{\text{KD}} = \mathbb{E}_{z,y}\left[\|G_t(z,y) - G_s(z,y)\|_2^2\right] \tag{5}$$

Finally, we update the generator loss from Equation (1) to incorporate this term:

Equation (6): Final generator loss with KD

$$\mathcal{L}_G = \mathcal{L}_{\text{cDCGAN}} + \lambda_{\text{KD}} \cdot \mathcal{L}_{\text{KD}} \tag{6}$$

Figure 7 depicts our proposed method.

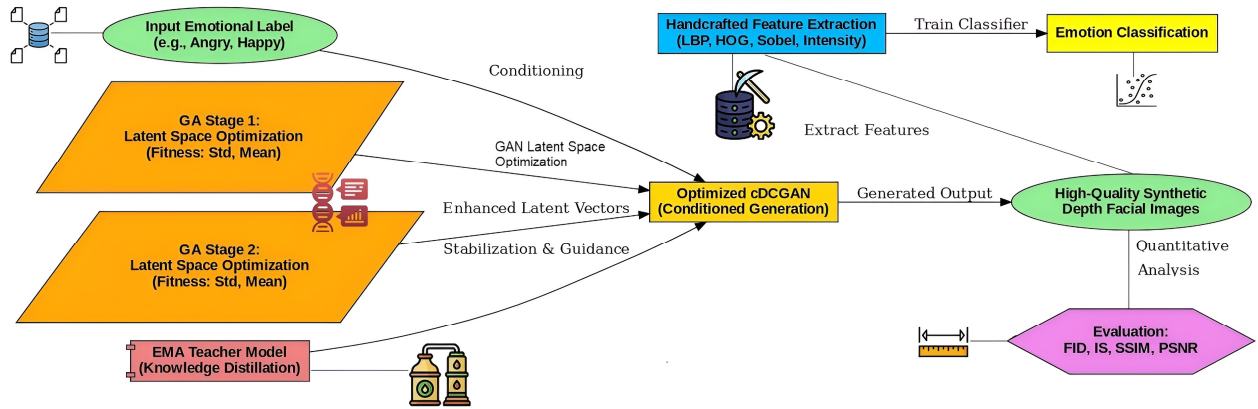

Figure 7. Our proposed synthetic data generation method

- **Evaluation and Results**
  - **The Dataset**

The Iranian Kinect Face Database (IKFDB)[7] [16] is a comprehensive RGB-D facial expression dataset collected using a Microsoft Kinect v2 sensor, which we use in this paper. It includes over 100,000 synchronized color and depth frames from 40 Iranian subjects (24 males and 16 females), ranging in age from 7 to 70 years. The dataset captures both RGB images at a resolution of 1920×1080 (cropped to 256×400) and depth images at 512×424 (cropped to 128×200), making it particularly valuable for depth-based facial analysis. Each participant performed seven basic facial expressions—neutral, happiness, anger, sadness, surprise, disgust, and fear across multiple video sequences, with each sequence containing 150–250 frames recorded at 30 frames per second. The first 50 frames of each sequence focus on subtle micro-expressions, while the remaining frames display fully developed emotional expressions. IKFDB fills a critical gap in facial datasets by representing Middle Eastern facial structures and offering high-quality depth data, varying head poses, and a broad age range. It has been evaluated with several classification models, yielding high Facial Expression Recognition (FER) accuracies: 90% with SVM, 92% with Multi-Layer Neural Networks (MLNN), and 95% with CNN. For Facial Micro-Expression Recognition (FMER), the models achieved 72%, 76%, and 85% accuracy, respectively. Face recognition rates were also impressive, reaching 99.2% with SVM, 99.7% with MLNN, and 100% with CNN, with 100% accuracy in face detection. The inclusion of detailed, high-resolution depth imagery sets IKFDB apart as a powerful resource for advancing research in facial expression analysis, emotion recognition, and biometric systems. Some depth samples of this dataset are illustrated in Figure 8. We employed three emotions of neutral, happy, and fear in this paper.

---
[7] https://link.springer.com/article/10.1007/s42452-020-03999-y

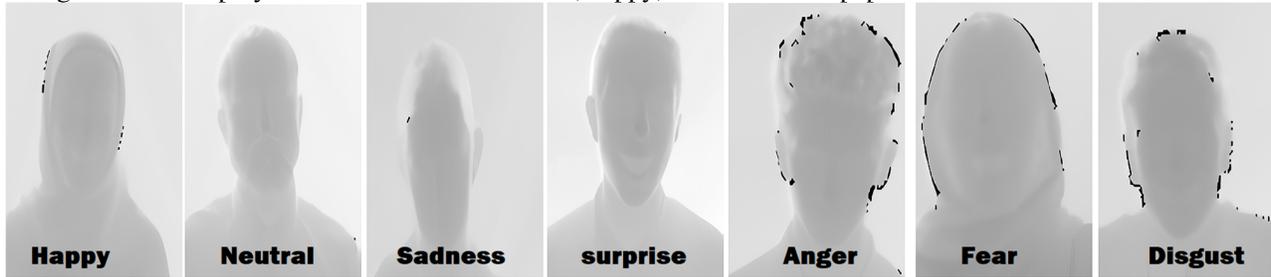

Figure 8. Selected depth samples of the IKFDB dataset

> **Data Processing, Features, and Classifiers**

Initially, each depth image is resized to a uniform resolution (256 by 256) to standardize the input space for subsequent feature extraction. Given the structural richness of depth data, we employed multiple complementary feature descriptors to capture both texture and geometry-based information from facial regions. First, HOG features are extracted to encode the local shape and gradient structures, which are particularly effective in representing facial muscle deformations and contours. Second, LBP histograms are computed to capture micro-texture details in the depth maps, such as subtle variations around facial landmarks. The LBP method works by comparing each pixel to its neighboring pixels and encoding this relationship into binary patterns, making it robust to illumination variations and particularly useful in facial analysis tasks. Third, we apply Sobel filtering to extract edge-based features that emphasize boundaries and transitions within the depth surface, offering cues about facial geometry and expressions. Additionally, intensity-based histograms of the grayscale depth images are calculated to provide global statistical information about pixel intensity distributions. These four types of features, HOG, LBP, Sobel, and intensity histograms, are concatenated (fused) into a single composite feature vector for each image, enhancing the representation power by incorporating both local and global depth characteristics. Following feature extraction, the data is partitioned into training and testing sets to facilitate model evaluation under controlled conditions. We then evaluate the effectiveness of three widely used classifiers: Random Forest (RF) [60], Decision Tree (DT) [61], and XGBoost (XGB) [62]. These models are chosen for their interpretability, robustness to noise, and effectiveness in handling high-dimensional feature spaces. Each classifier is trained on the extracted feature vectors and evaluated using standard classification metrics. This modular and interpretable framework enables reliable recognition of facial emotions from depth images and is well-suited for real-time or embedded applications in affective computing, human-computer interaction, and assistive technologies. Figure 9 presents KDE plots of the first Linear Discriminant (LDA) [69] component distributions across our three emotion classes (Fear, Happiness, Neutral) for three data categories: baseline, synthetic (the proposed method), and baseline plus synthetic. The LDA is a supervised dimensionality reduction technique that projects data onto a lower-dimensional space while maximizing class separability. Here, LDA was used to combine and compress all extracted features into a single discriminative component (LDA1) to visualize how well the emotion classes are separated. In the baseline plot (left), the classes are already distinguishable, with fear separated, though happiness and neutral have some overlap. The synthetic plot (middle) shows slightly better separation between happiness and neutral, showing that the synthetic data helps diversify inter-class boundaries. In the baseline plus synthetic plot (right), the combination results in even more distinct and well-separated distributions, particularly sharpening the boundary between all three classes. Overall, the fusion of synthetic data with real samples improves class separability in the LDA space, which can enhance classification performance.

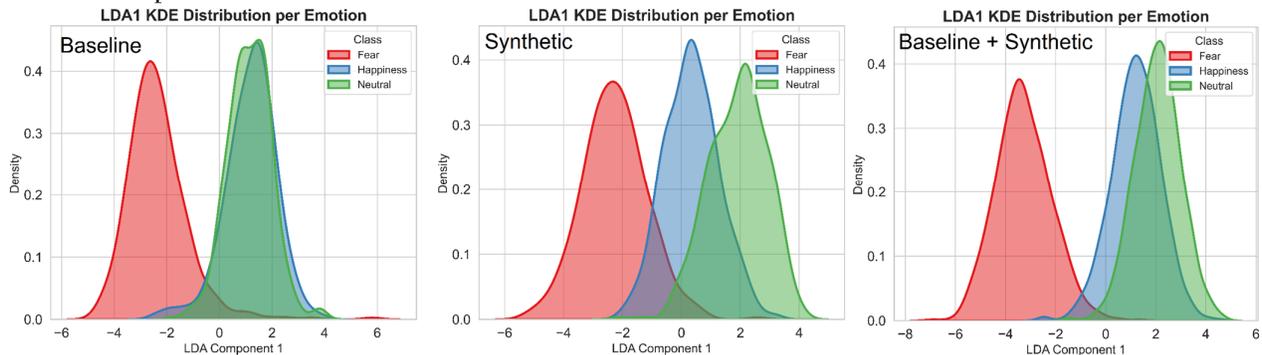

Figure 9. LDA-based KDE distribution of emotion classes across all data combinations (proposed method)

➢ **SDG Evaluation Metrics**

To quantitatively evaluate and compare the perceptual and structural quality of the synthetic depth images generated by our method, we employed a set of well-established image quality assessment metrics: FID, IS, SSIM, and PSNR. These metrics collectively provide a comprehensive evaluation framework by capturing both perceptual fidelity and pixel-level accuracy. The FID measures the distance between feature distributions of real and generated images using the activations of a pre-trained network, typically Inception-v3[8]. A lower FID score shows that the synthetic images are closer to the real distribution, which is crucial for the realism of depth facial features. The IS evaluates both the quality and diversity of generated samples by computing the Kullback–Leibler (KL) divergence [63] between the conditional label distribution and the marginal label distribution predicted by a classifier. The KL divergence is a foundational measure in information theory and statistics that quantifies how one probability distribution diverges from another. Higher IS values suggest that the generated images are not only realistic but also varied, which is important for capturing a wide range of emotional expressions in depth format. The SSIM, on the other hand, focuses on structural information by comparing luminance, contrast, and structural similarity between the original and generated depth images. This metric is particularly important in our context as it evaluates whether the generated images preserve facial geometry and contour integrity. Finally, the PSNR quantifies the pixel-wise reconstruction fidelity by measuring the ratio between the maximum possible power of a signal and the power of corrupting noise, making it useful for assessing image degradation. Higher PSNR values indicate better preservation of the original image structure. Together, these metrics were used to rigorously assess and compare the quality of our generated depth images against baseline or state-of-the-art methods, demonstrating the effectiveness and realism of our synthetic data.

➢ **Experiments and Results**

All SDG simulations have been conducted on by CUDA on an RTX 3050 (6 GB) graphics card, and it tripled the computation time by three times. All deep learning algorithms run over 500 epochs, 256 latent dimensions, and a 300 batch size in general. Regarding our method generator learning rate was 0.00008, and the discriminator learning rate was 0.00001. Furthermore, the KD parameters were EMA decay value = 0.994, lambda generation value = 1.0, and lambda discrimination value = 0.5. Finally, GA parameters were population size = 20, number of generations = 100, crossover probability = 0.8, mutation probability = 0.2. We compared our method results with the baseline, VAE, GAN, GMM, and KDE for all metrics. Figure 10 represents the 2D t-distributed Stochastic Neighbor Embedding (t-SNE) [70] projections of our extracted features for our three emotions of fear, happiness, and neutral for three sets of baseline, synthetic, and baseline plus synthetic. The t-SNE is a nonlinear dimensionality reduction technique that maps high-dimensional data into a low-dimensional space, preserving local similarities and cluster structures. It is widely used for visualizing complex data distributions such as embeddings or feature vectors. In the baseline (left), data points from different classes are somehow mixed, indicating normal separability. The synthetic data (middle) shows improved clustering and clearer boundaries among classes, suggesting the synthetic data carries discriminative structure. The baseline plus synthetic plot (right) demonstrates even tighter class groupings, especially for happiness and neutral, confirming that augmentation boosts feature space structure.

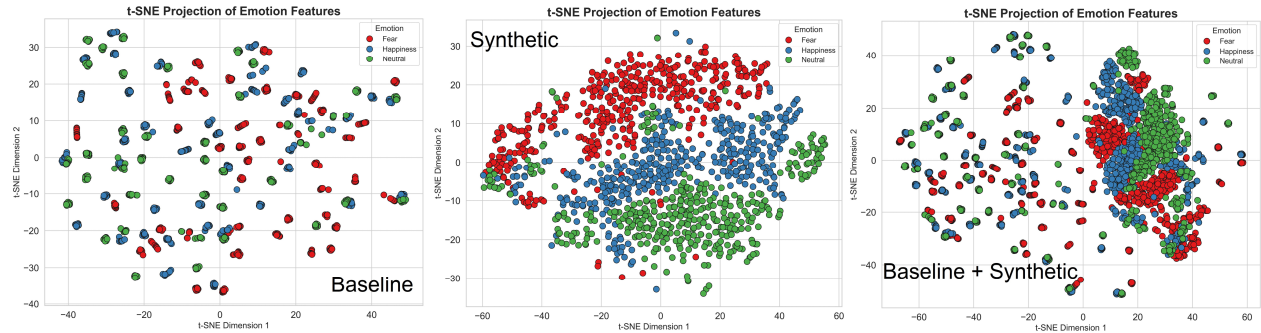

Figure 10. t-SNE visualization of emotion feature distributions across baseline, synthetic, and combined sets (proposed method)

Figure 11 illustrates the correlation heatmaps between our four extracted feature types, HOG, LBP, Sobel edge, and intensity histogram, across our three sets. In the baseline (left), HOG, LBP, and intensity histogram features are highly correlated (>0.8), indicating strong redundancy, while Sobel edge is weakly or negatively correlated with the others. In the synthetic data (middle), correlations are more varied and less extreme, suggesting a more diverse feature space with reduced redundancy. The combined heatmap (right) shows overall lower correlations, especially for HOG and Sobel, showing improved feature diversity. This suggests that synthetic data helps decorrelate features, potentially improving classification performance.

---
[8] https://keras.io/api/applications/inceptionv3/

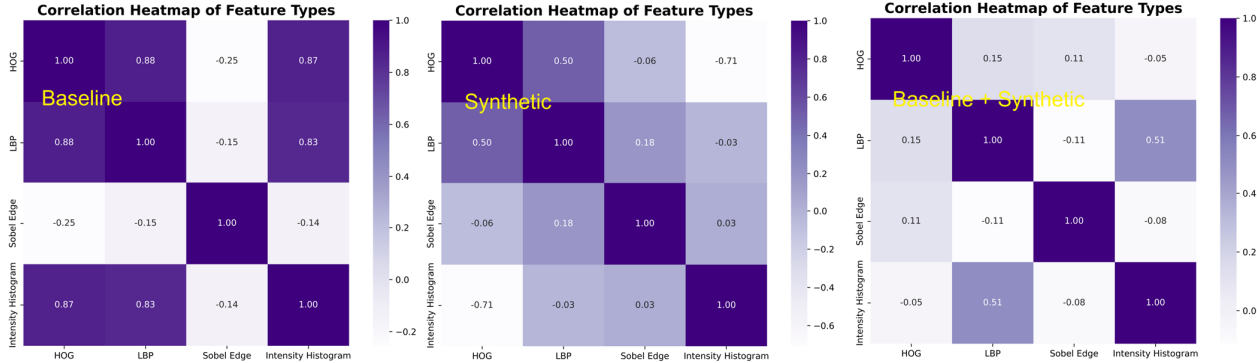

Figure 11. Correlation heatmaps of features across baseline, synthetic, and combined sets (proposed method)

Figure 12 shows the training progress of the proposed method over 500 epochs. The Discriminator Loss (D) and Generator Loss (G) fluctuate initially but gradually stabilize, representing adversarial training convergence. The Generator Knowledge Distillation Loss (green dashed line) remains steady, showing consistent alignment with the teacher model's knowledge. Similarly, the Discriminator KD Loss (red dashed line) stays low, showing effective learning without destabilizing the GAN. So, the graph reflects a balanced co-training between adversarial objectives and distillation, supporting the stability and effectiveness of the proposed method.

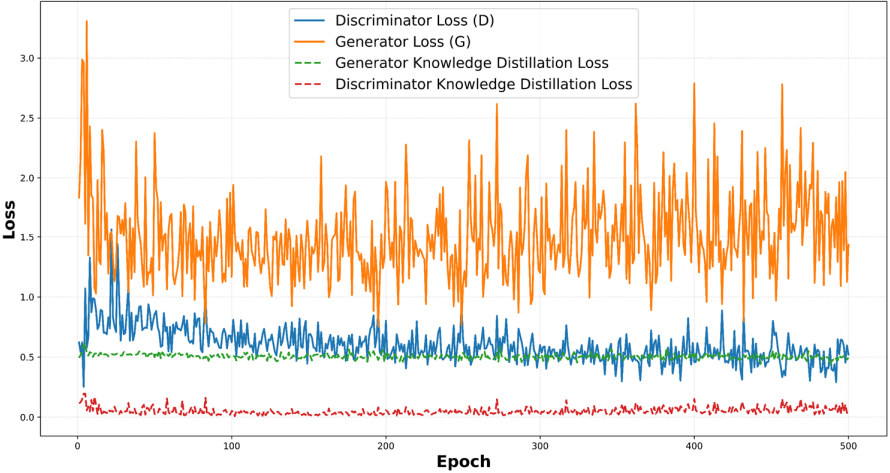

Figure 12. Training loss curves of our knowledge distilled cDCGAN

Table 1 presents a detailed comparison of classification performance using different synthetic data generation methods, VAE, GAN, GMM, KDE, and the proposed method, evaluated across three classifiers: Random Forest (RF), Decision Tree (DT), and XGBoost (XGB), under both synthetic-only and synthetic plus original (augmented) scenarios. The baseline, trained solely on original data, yields excellent results with Accuracy, Precision, Recall, and F1 Score all around 97.8%, serving as an upper-bound reference. When using only synthetic data (1200 samples), the proposed method consistently outperforms others across all classifiers: RF reaches 91.67% accuracy, XGB achieves 94.58%, and DT improves to 58.92%, surpassing other generators like VAE, GAN, GMM, and KDE, which often show lower accuracy and less balanced metrics. In the augmented (synthetic + original) setup with 2400 samples, performance improves further, RF reaches 92.50%, DT 68.92%, and XGB peaks at 95.46%, again with the proposed method outperforming all others in nearly every metric. Notably, GMM and KDE consistently lag behind across all scenarios, suggesting they produce less effective or noisier synthetic samples. Overall, the table demonstrates that the proposed method generates high-quality, class-consistent synthetic data that not only performs well on its own but also enhances classifier performance when combined with real data, leading to improved generalization, robustness, and higher reliability across all tested models. Also, Figure 13 depicts the same results as a line plot.

Table 1. Classification Results (80% Train and 20% Test)

| Baseline | Classifier | VAE | GAN | GMM | KDE | Proposed |
|---|---|---|---|---|---|---|
| Accuracy: 97.92%<br>Precision: 97.94%<br>Recall: 97.79%<br>F1 Score: 97.84% | RF (Synthetic) ↑<br>1200 Samples<br>400 Per Class | Accuracy: 81.67%<br>Precision: 81.39%<br>Recall: 81.39%<br>F1 Score: 81.39% | Accuracy: 85.42%<br>Precision: 84.66%<br>Recall: 84.70%<br>F1 Score: 84.66% | Accuracy: 65.42%<br>Precision: 64.55%<br>Recall: 64.66%<br>F1 Score: 64.52% | Accuracy: 60.42%<br>Precision: 59.22%<br>Recall: 59.48%<br>F1 Score: 59.26% | **Accuracy: 91.67%**<br>**Precision: 91.56%**<br>**Recall: 91.62%**<br>**F1 Score: 91.57%** |
| -<br>-<br>-<br>- | RF (Syn + Org) ↑<br>2400 Samples<br>800 Per Class | Accuracy: 90.21%<br>Precision: 90.21%<br>Recall: 90.16%<br>F1 Score: 90.15% | Accuracy: 90.42%<br>Precision: 90.32%<br>Recall: 90.21%<br>F1 Score: 90.19% | Accuracy: 77.08%<br>Precision: 77.10%<br>Recall: 76.95%<br>F1 Score: 76.95% | Accuracy: 77.50%<br>Precision: 77.27%<br>Recall: 77.29%<br>F1 Score: 77.26% | **Accuracy: 92.50%**<br>**Precision: 92.46%**<br>**Recall: 92.49%**<br>**F1 Score: 92.46%** |
| Accuracy: 83.33%<br>Precision: 83.23%<br>Recall: 83.21%<br>F1 Score: 83.21% | DT (Synthetic) ↑<br>1200 Samples<br>400 per class | Accuracy: 52.92%<br>Precision: 53.13%<br>Recall: 52.81%<br>F1 Score: 52.87% | Accuracy: 57.50%<br>Precision: 56.23%<br>Recall: 56.57%<br>F1 Score: 56.37% | Accuracy: 51.50%<br>Precision: 52.38%<br>Recall: 51.83%<br>F1 Score: 52.04% | Accuracy: 46.25%<br>Precision: 46.06%<br>Recall: 45.66%<br>F1 Score: 45.77% | **Accuracy: 58.92%**<br>**Precision: 58.22%**<br>**Recall: 57.95%**<br>**F1 Score: 57.87%** |
| -<br>-<br>-<br>- | DT (Syn + Org) ↑<br>2400 Samples<br>800 Per Class | Accuracy: 67.50%<br>Precision: 67.85%<br>Recall: 67.48%<br>F1 Score: 67.51% | Accuracy: 66.42%<br>Precision: 66.06%<br>Recall: 65.33%<br>F1 Score: 65.43% | Accuracy: 65.21%<br>Precision: 65.15%<br>Recall: 65.07%<br>F1 Score: 65.09% | Accuracy: 60.00%<br>Precision: 59.94%<br>Recall: 59.78%<br>F1 Score: 59.80% | **Accuracy: 68.92%**<br>**Precision: 68.75%**<br>**Recall: 67.76%**<br>**F1 Score: 67.75%** |
| Accuracy: 97.92%<br>Precision: 97.88%<br>Recall: 97.90%<br>F1 Score: 97.89% | XGB (Synthetic) ↑<br>1200 Samples<br>400 per class | Accuracy: 86.25%<br>Precision: 86.44%<br>Recall: 85.66%<br>F1 Score: 85.79% | Accuracy: 88.75%<br>Precision: 88.17%<br>Recall: 88.19%<br>F1 Score: 88.16% | Accuracy: 69.58%<br>Precision: 68.69%<br>Recall: 68.71%<br>F1 Score: 68.48% | Accuracy: 61.67%<br>Precision: 60.95%<br>Recall: 61.01%<br>F1 Score: 60.88% | **Accuracy: 94.58%**<br>**Precision: 94.47%**<br>**Recall: 94.58%**<br>**F1 Score: 94.49%** |
| -<br>-<br>-<br>- | XGB (Syn + Org) ↑<br>2400 Samples<br>800 Per Class | Accuracy: 93.96%<br>Precision: 93.93%<br>Recall: 93.97%<br>F1 Score: 93.94% | Accuracy: 94.17%<br>Precision: 94.03%<br>Recall: 94.03%<br>F1 Score: 94.03% | Accuracy: 81.25%<br>Precision: 81.24%<br>Recall: 81.10%<br>F1 Score: 81.14% | Accuracy: 76.67%<br>Precision: 76.49%<br>Recall: 76.46%<br>F1 Score: 76.47% | **Accuracy: 95.46%**<br>**Precision: 96.48%**<br>**Recall: 96.43%**<br>**F1 Score: 96.42%** |

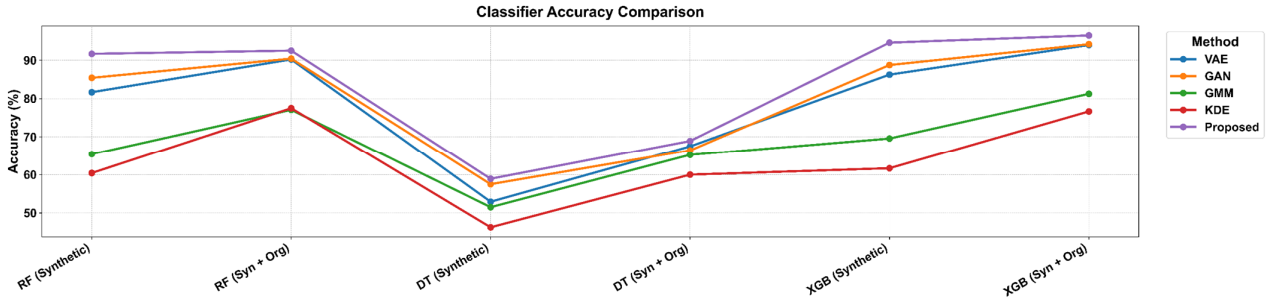

Figure 13. Classification results as a line plot based on Table 1

Table 2 compares the quality of synthetic depth facial images generated by different methods, VAE, GAN, GMM, KDE, and the Proposed method, using four standard image evaluation metrics: FID (Fréchet Inception Distance), IS (Inception Score), SSIM (Structural Similarity Index Measure), and PSNR (Peak Signal-to-Noise Ratio). Lower FID values indicate higher similarity to real images, and here the Proposed method achieves the best score (45.56), significantly outperforming others like GAN (129.84), VAE (153.56), GMM (166.36), and KDE (179.26), showing its superior ability to generate realistic samples. For IS, which reflects both image quality and diversity, the proposed method again leads (4.14 ± 0.17), surpassing the baseline (3.57 ± 0.16) and all other synthetic methods, indicating that its outputs are both sharp and varied. In terms of SSIM, which measures structural similarity, the proposed method scores 0.71, noticeably higher than GAN (0.60), VAE (0.58), and much better than GMM (0.46) and KDE (0.35), demonstrating better preservation of facial structure. Finally, for PSNR, which evaluates image clarity and noise level, the proposed method achieves 20.68 dB, showing cleaner and sharper images than all other methods, with KDE and GMM performing the worst. In total, across all four metrics, the proposed method significantly outperforms existing approaches, producing synthetic depth images that are more realistic, diverse, and structurally faithful. The bar plot of Table 2 is illustrated in Figure 14.

Table 2. Synthetic Data Evaluation Results

| Baseline | | VAE | GAN | GMM | KDE | Proposed |
|---|---|---|---|---|---|---|
| - | FID ↓ | 153.5658 | 129.8420 | 166.3631 | 179.2621 | **45.5560** |
| 3.5695 ± 0.1608 | IS ↑ | 2.8474 ± 0.0811 | 3.2655 ± 0.2122 | 2.1022 ± 0.2720 | 2.1381 ± 0.1842 | **4.1447 ± 0.1659** |
| - | SSIM ↑ | 0.58 | 0.60 | 0.46 | 0.35 | **0.71** |
| - | PSNR ↑ | 15.93 dB | 15.59 dB | 13.60 dB | 10.36 dB | **20.68 dB** |

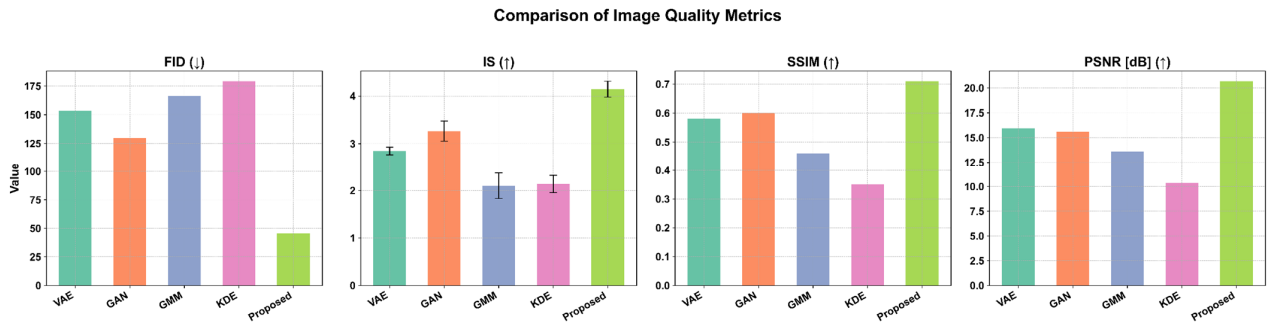

Figure 14. Synthetic data evaluation results from comparison based on Table 2

Figure 15 presents the GA loss progression of our method over 100 iterations for three emotion classes: Neutral, Happiness, and Fear. Each subplot shows the fitness improvement (interpreted here as increasing "loss" due to a maximization objective) as the GA evolves better synthetic samples per class. For all three classes, the curves exhibit a steep rise in the early iterations, followed by a plateau, indicating that the algorithm quickly discovers significantly better individuals early on and then gradually fine-tunes them. The neutral and happiness classes reach a fitness value around 9.0–9.1, while the fear class converges slightly lower, around 8.8, though with a similarly smooth progression. These consistent upward trends show that the GA successfully optimizes class-specific generation, progressively improving the quality of synthesized samples with respect to the objective function (here fitness) used.

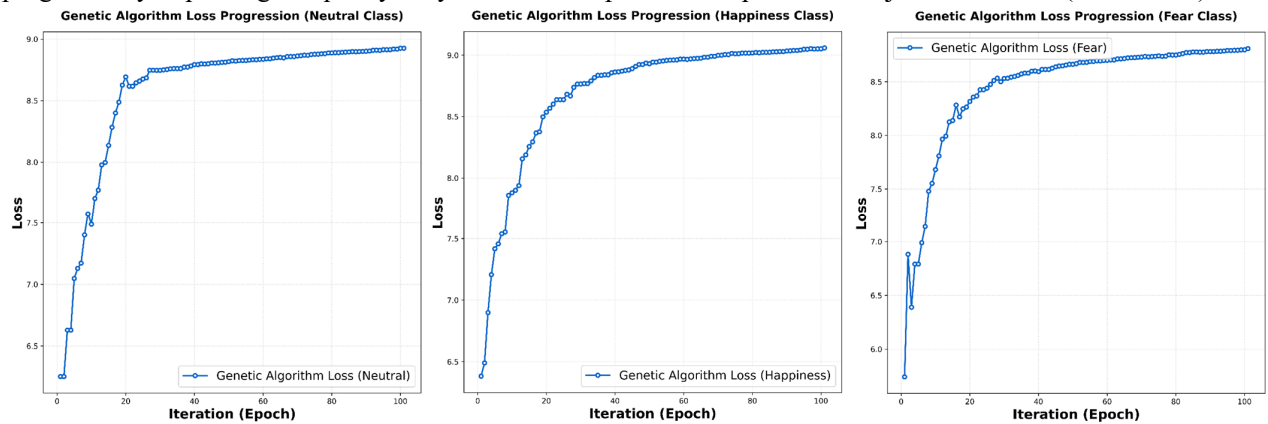

Figure 15. Genetic algorithm loss progression for class-specific synthetic sample optimization

➢ **Discussion**

In this work, we address the critical need for rich and diverse depth facial data in affective computing by proposing a comprehensive synthetic data generation framework designed for emotion-specific synthesis. To tackle the first question of RQs raised in the introduction section, we demonstrate that our method effectively mitigates dataset scarcity by generating high-quality, class-consistent depth face images that significantly enhance classifier performance when used alone or in combination with real data. Addressing the second question, we integrate Knowledge Distillation (KD) using an Exponential Moving Average (EMA) teacher model, which stabilizes GAN training, reduces mode collapse, and improves image consistency across epochs. For the third question, we extend a Genetic Algorithm (GA) into the GAN latent space, evolving class-specific latent vectors that produce emotionally diverse and visually robust depth images, as shown in our per-class optimization curves. In response to the fourth question, we extract traditional texture and structure-based features (LBP, HOG, and Sobel) from the generated images and verify that they preserve discriminative emotional cues, enabling reliable emotion classification across multiple machine learning models. Finally, to address the fifth question, we evaluate our synthetic images using standard quantitative metrics—FID, IS, SSIM, and PSNR—and consistently outperform alternative generation methods such as VAE, GMM, and KDE, confirming both the visual fidelity and emotional diversity of our samples. Through these contributions, we present a robust and interpretable pipeline that advances synthetic depth face generation for emotion recognition. Figure 16 visually compares depth face samples generated by different methods: VAE, GAN, GMM, KDE, baseline (real), and our proposed method. The Proposed method produces the most realistic and artifact-free depth images, closely matching the quality and structure of the baseline samples. In contrast, other methods show significant issues such as blurriness (VAE), deformation (GAN, GMM), or extreme noise and inversion artifacts (KDE).

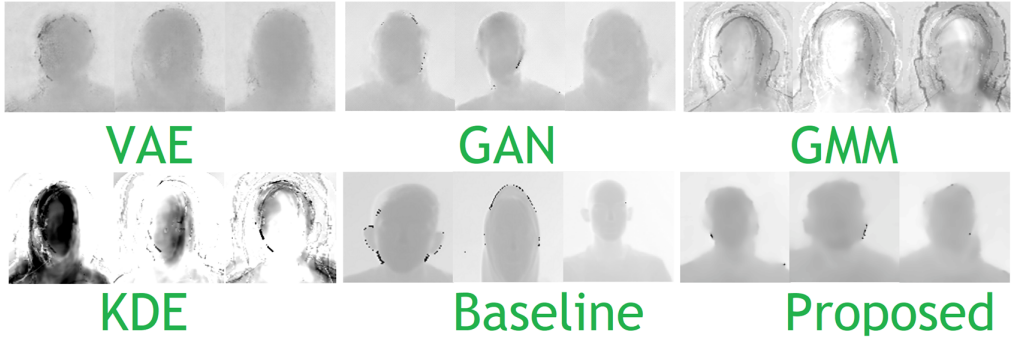

Figure 16. Visual comparison of depth face samples generated by different methods

- **Conclusion**

In this study, we addressed the longstanding challenge of limited and emotion-specific depth facial datasets in affective computing by introducing a novel, hybrid synthetic data generation framework. Our method integrates three core components: a conditional Deep Convolutional GAN (cDCGAN) for emotion-conditioned depth face synthesis, an Exponential Moving Average (EMA)-based Knowledge Distillation (KD) mechanism to stabilize and guide training, and a Genetic Algorithm (GA) that evolves latent vectors post-training to enhance visual diversity and emotional distinctiveness. This combination enables us to generate high-fidelity, structurally coherent, and semantically consistent depth images designed to target emotions. Extensive experiments across multiple axes prove the strength of our approach. Quantitatively, we outperform other generative models, including GAN, VAE, GMM, and KDE, in all quality metrics: FID (↓45.56), IS (↑4.14), SSIM (↑0.71), and PSNR (↑20.68 dB). Qualitatively, visual comparisons show that our proposed method generates synthetic depth faces that closely resemble real samples with minimal noise or artifacts. Moreover, emotion classification results using handcrafted features (LBP, HOG, Sobel, and intensity histograms) and standard classifiers (RF, DT, XGBoost) further confirm the utility of our generated samples in downstream tasks. Notably, our synthetic data combined with real samples significantly boosts performance—XGBoost reaches up to 96% recognition accuracy, showing our method's effectiveness in both data augmentation and standalone synthesis contexts.

> **Future Works**

While our current method shows impressive gains in quality and performance, future work could explore integrating more advanced generative models, such as diffusion models or transformer-based generators, to further push the boundaries of realism and diversity. Additionally, using emotion-aware loss functions and perceptual metrics during GAN training may offer finer control over subtle affective expressions. Extending the approach to multi-view or temporal depth data could also open up new applications in video-based emotion recognition and behavioral analysis.

> **Suggestions**

We recommend that future research in affective computing adopt similar hybrid frameworks that combine generative learning with evolutionary strategies and knowledge transfer, as they offer a powerful balance between quality, control, and interpretability. Furthermore, using both visual (qualitative) and quantitative metrics side-by-side, as we have done through FID, IS, PSNR, SSIM, and classification performance, can provide a more holistic evaluation of synthetic data utility in real-world emotion recognition tasks.

- **Annex**

Annex 1 illustrates the distribution of random forest classifier performance across four key evaluation metrics, accuracy, precision, recall, and F1-score, for three different training scenarios: baseline, synthetic, and baseline plus dsynthetic, based on 30 independent runs each. In the baseline setup (top row), all metric distributions are sharply peaked between 0.98 and 1.00, with minimal variance, indicating extremely stable and high performance. However, such tight clustering may reflect overfitting, as the model likely memorizes patterns from a limited dataset. In the synthetic setup (middle row), where the model is trained only on data generated by our framework, the performance drops to around 0.88–0.90, and the violins widen significantly. This reveals higher variability and broader generalization, representing the increased diversity introduced by the synthetic data. Finally, the baseline plus synthetic condition (bottom row) presents a balanced improvement, with scores consistently ranging from 0.91 to 0.94 and moderate variance, showing that the hybrid dataset helps reduce overfitting while still maintaining high accuracy. Notably, F1-score and recall in this setup show both stability and improvement, implying better class balance and generalization. Therefore, these violin plots based on 30 runs support that combining synthetic and real data enhances both robustness and reliability of the classifier in a statistically consistent manner.

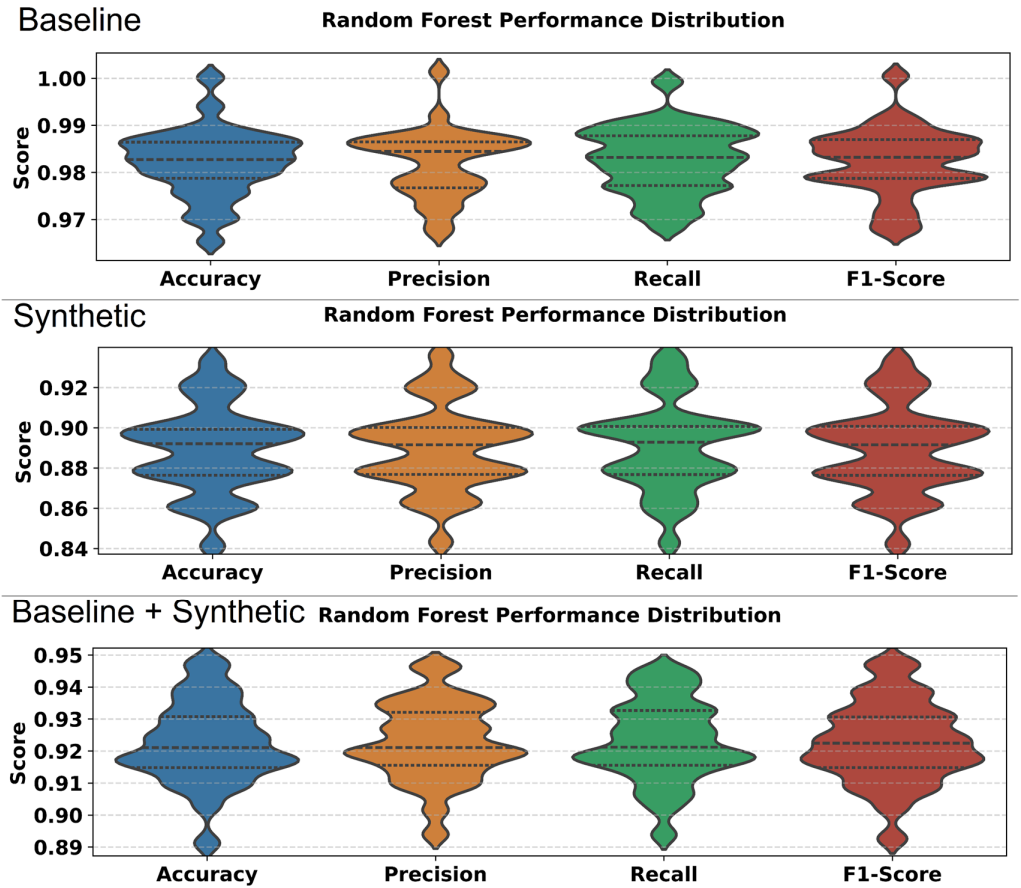

Annex 1. *Violin plots showing the distribution of random forest classification performance (accuracy, precision, recall, F1-score) across 30 runs for baseline, synthetic, and combined sets.*

### Ethical and Conflict of Interest Statement

All depth facial images used in this study are either synthetically generated or collected by the authors in a separate research project. Depth images inherently lack identifiable texture, color, or fine-grained appearance information, making them non-identifiable and anonymized by nature. The dataset used in this work does not contain any personally recognizable features, and no individual identities can be inferred from the data. As such, no ethical approval or informed consent was required. This research was conducted entirely without external funding and is fully self-funded by the authors. The study was carried out independently by the authors over a six-month period, from early February to the end of July 2025. The authors declare that there are no conflicts of interest, financial or otherwise, related to the content or outcomes of this work.

### Code and Data Availability

Link to the Data on Kaggle: https://kaggle.com/datasets/5382fad0083f66167d9191f9392806f8b3d4f856b7b0ce11f3c71ed8545ccfb9

Link to the Code on GitHub: https://github.com/SeyedMuhammadHosseinMousavi/Synthetic-Data-Generation-for-Emotional-Depth-Faces

### Roles and Responsibilities

Author 1 (Seyed Muhammad Hossein Mousavi) was responsible for the majority of the work, including conceptualization, methodology design, software implementation, data curation, formal analysis, visualization, writing the original draft, and conducting the majority of experiments and evaluations. Author 2 (S. Younes Mirinezhad) contributed to writing – review & editing, provided support in methodological refinement, and assisted with interpretation of results and manuscript proofreading. Both authors reviewed and approved the final manuscript, with Author 1 carrying out the primary technical and research workload.